\documentclass{article}
\usepackage{spconf,amsmath,graphicx}
\usepackage{todonotes}
\usepackage{caption}
\usepackage{subcaption}
\usepackage{dsfont}
\usepackage{hyperref}
\hypersetup{
    colorlinks=true,
    linkcolor=blue,
    filecolor=magenta,      
    urlcolor=cyan,
}

\title{End-to-end detection-segmentation network with ROI convolution}
\name{Zichen Zhang$^{\star}$, Min Tang$^{\star}$, Dana Cobzas$^{\star}$, Dornoosh Zonoobi$^{\dagger}$, Martin Jagersand$^{\star}$, Jacob L. Jaremko$^{\dagger}$ }
\address{ $^{\star}$ Computing Science Department, University of Alberta, Edmonton, Canada \\
    $^{\dagger}$ Radiology Department, University of Alberta, Edmonton, Canada}

\begin{document}
%
\maketitle
\begin{abstract}
We propose an end-to-end neural network that improves the segmentation accuracy of fully convolutional networks by incorporating a localization unit. This network performs object localization first, which is then used as a cue to guide the training of the segmentation network.
We test the proposed method on a segmentation task of small objects on a clinical dataset of ultrasound images. We show that by jointly learning for detection and segmentation, the proposed network is able to improve the segmentation accuracy compared to only learning for segmentation.
Code is publicly available at \url{https://github.com/vincentzhang/roi-fcn}.
\end{abstract}
\section{Introduction}
\label{sec:intro}
Fully Convolutional Network (FCN) \cite{fcn} 
is the de-facto network architecture
in semantic segmentation and has been the basis for many advanced network
models.
%
U-Net \cite{unet2d} was developed specifically for medical imaging segmentation. The network had a U shape architecture where it tried to make upsampling and downsampling streams symmetrical, with skip connections that concatenated the feature maps from downsampling layers to those in the upsampling layers.
Simon et al. \cite{tiramisu} employed dense skip connections for semantic segmentation. Similar to U-Net, their network was also in a U shape. 
The dense skip connections were added to the downsampling stream only. 
The common architectural design is that the input image will go through downsampling layers (convolution and pooling). Then the encoded features will be passed through upsampling layers (transposed convolution) to get the score map of the same size as the input.
All these networks are trained with the same cross-entropy loss function that essentially makes the network learn a classifier for each pixel and average or sum up the loss for all pixels.
This makes it hard to train the network when the object to segment is small. 

Intuitively, humans perform semantic segmentation by first attending to some rough region where each pixel inside that region is then classified. In other words, object detection can be a first step for accurate semantic segmentation.
There have been some attempts in using object detection for segmentation tasks \cite{boxsup, masking}, but they relied on external region proposal methods such as Selective Search \cite{selective}.

Some recent works have started to jointly learn detection and segmentation \cite{blitznet, maskrcnn}. 
Mask RCNN \cite{maskrcnn} is the state-of-the-art for instance segmentation where it first predicts the bounding box of each object instance in the image, then predicts pixel-wise labels inside each predicted box. The network was trained end-to-end with the multi-task loss of localization, classification and segmentation. It was shown that this multi-task loss improved the accuracy compared to training for each task individually. 
However, a pooling layer called "ROI-align" was used to transform the regions of interest (ROI) to fixed size input for the segmentation stream. It reduced the resolution of the features which caused a loss of spatial details especially along the object boundary. Also the segmentation stream caused computational overhead since it processed each region individually.
BlitzNet \cite{blitznet} adopted a fully convolutional architecture and performed image-wise computation. But it did not make explicit use of the detection result to improve segmentation accuracy. 

\begin{figure*}[tbp]
\begin{center}
   \includegraphics[trim=0 2cm 0 1cm, clip=True, width=0.6\linewidth]{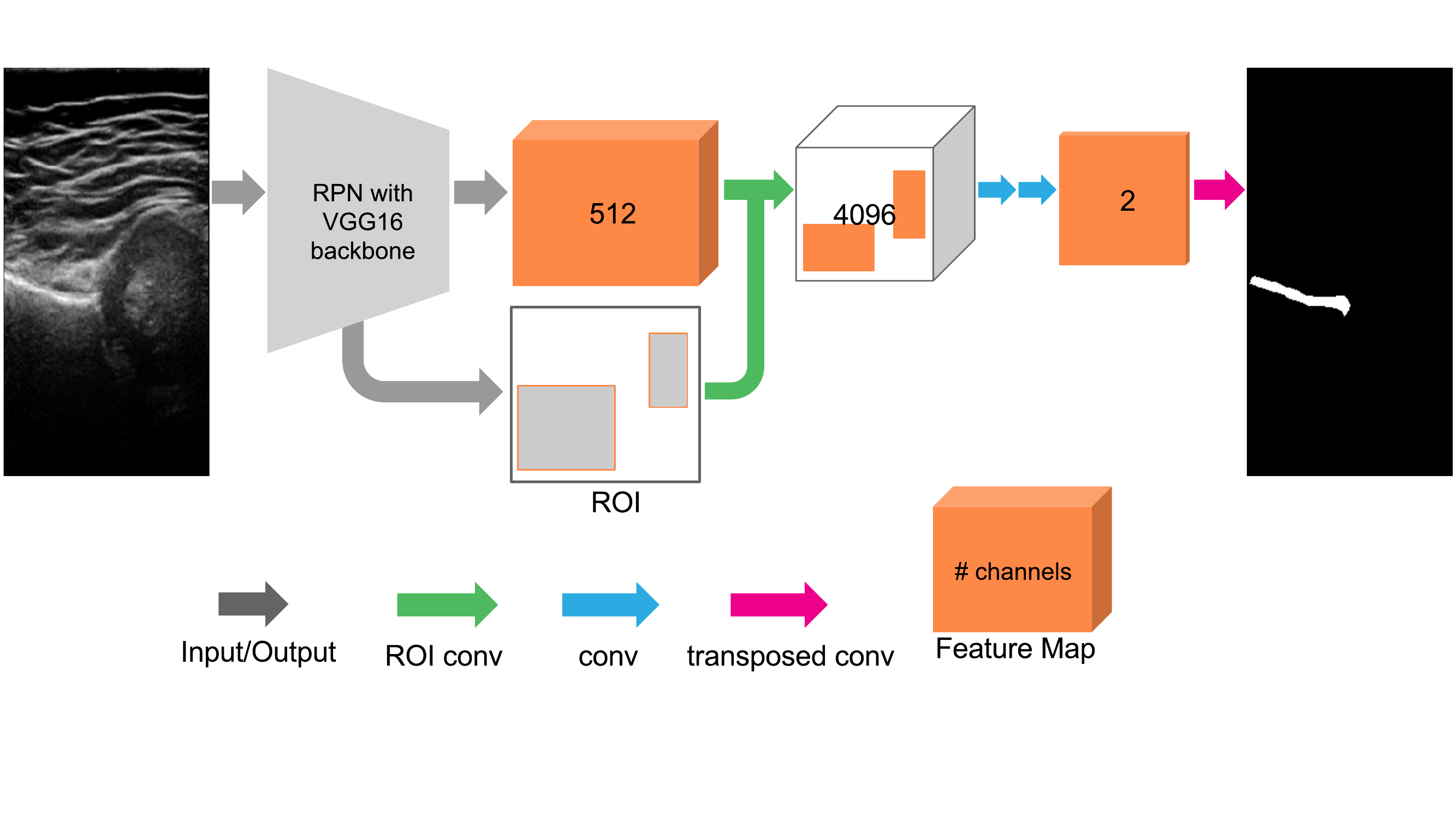}
   \vspace{-0.7cm}
\end{center}
   \caption{Overall architecture of our network. The legend at the bottom illustrates different components. Best viewed in color. The input to the network can be any 2D grayscale (duplicated to 3 channels) or color image. The RPN network is built on top of VGG16. It takes an input image and outputs feature maps of 512 channels and a number of ROIs. The arrow for `ROI conv' represents one ROI convolutional layer that takes as input the feature map and the ROIs. Each `conv' and `transposed conv' arrow represents one convolutional layer and transposed convolutional layer, respectively.}
\label{fig:workflow}
\end{figure*}

In this paper, we propose a fully convolutional architecture that incorporates an object localization unit to improve semantic segmentation accuracy. 
Our work addresses the issues discussed above by introducing a new type of convolutional layer named ROI convolution, that replaces the ROI pooling layer by convolving directly on top of the region proposals from the localization network. It applies convolution on the features inside all the ROIs in one shot without the need of passing them individually through the downstream convolutional layers. 
This proposed network performs image-wise computation and can be trained end-to-end. 
To demonstrate the efficacy of our method, we apply it to the problem of segmentation of acetabulum from ultrasound images. It is a challenging task since the object of interest is small. We show that with the localization unit, our method performs well despite the  high class imbalance.
%

\section{Proposed method}
\label{sec:format}
The overall architecture of the proposed network is shown in Figure \ref{fig:workflow}. In this section, we will go over the details of each component.

\noindent
\textbf{Backbone} 
We use VGG16 as the feature extractor, i.e., the backbone of the architecture. 
The network that we propose could be built on top of a deeper backbone like ResNet \cite{resnet} to further improve accuracy. We adopt the simpler backbone VGG16 just to demonstrate how the new architecture improves segmentation with detection. 

\noindent
\textbf{Localization unit} 
We adopt the Region Proposal Network (RPN) from Faster R-CNN \cite{faster} to perform object localization. Each region is classified as foreground or background. The foreground regions are passed to the downstream segmentation network where each pixel inside the regions is classified. 

\noindent
\textbf{ROI convolution} 
The output of the RPN is a list of ROIs in the form of bounding boxes. The features of the ROIs can be extracted from the feature map of the previous layer and passed to downstream segmentation network. Region-wise method has been widely used \cite{maskrcnn, faster, masking} where each ROI is processed individually as a new input image. To obtain a fixed-size input, each ROI is first fed into a pooling layer which produces a fixed-size grid. The features for each cell in the grid are those of the pixel that has the maximum activation in that cell. 
In contrast, we propose an image-wise method to simplify the computation, where the downstream segmentation network processes all the ROIs in one shot. In order to do that, a new layer ROI convolution is introduced. A similar idea has been proposed in \cite{roiconv}. But they used ROI convolution only in the inference time to speed up computation. We show that ROI convolution is differentiable and is trainable through backward propagation. We use ROI convolution at both training and inference time.

\noindent
\textbf{Forward and back propagation of ROI convolution}\\
The forward pass of ROI convolution is shown in Eq. \ref{eq:forward}.
\begin{equation}
y_{i,j}^l = (w^l * z^{l-1})_{i,j} = \mathds{1} [(i,j) \in \cup \text{ROI}] \sum_{b=0}^{k-1} \sum_{a=0}^{k-1} w_{a,b}^l z^{l-1}_{i+a,j+b} \
\label{eq:forward}
\end{equation}
where $y_{i,j}^l$ represents the pre-activation output of layer $l$ at pixel $(i,j)$.
$z^l$ represents the activation of layer $l$.
$w^l$ is the filter at layer $l$, of size $k \times k$.
$*$ denotes the ROI convolution.
$\cup\, \text{ROI}$ denotes the union of the set of indices inside all the ROIs. The indicator function checks if the pixel $(i,j)$ is inside any ROI.
The forward pass is just like the regular convolution except that the pixels outside the ROIs are set to zeros.

The backpropagation of ROI convolution is derived as follows.
Suppose that we have the cost function E, and 
$\frac{\partial E}{\partial z^l} = \delta^l$ for the current layer $l$, we need to compute $\frac{\partial E}{\partial z^{l-1}}$ and  $\frac{\partial E}{\partial {w^l}}$:
\begin{align}
\frac{\partial E}{\partial z^{l-1}_{i,j}} &= 
\sum^{k-1}_{a=0} \sum^{k-1}_{b=0}
(\delta^{l}\,\sigma')_{i-a,j-b}
w^l_{a,b} \mathds{1} [(i-a,j-b) \in \cup \text{ROI}]\nonumber\\
\frac{\partial E}{\partial w^l_{a,b}} &= 
\sum^{W^{l}_o}_{j=0} \sum^{H^{l}_o}_{i=0}
\delta^l_{i,j} \sigma'_{i,j}
z^{l-1}_{i+a,j+b} \mathds{1} [(i,j) \in \cup \text{ROI}]
\end{align}
where $W^{l}_o, H^{l}_o$ denotes the width and height of the output of layer $l$. $\sigma'$ denotes the derivative of the activation function.
Unlike in regular convolution, the pixels outside the ROIs do not contribute to the gradient in the ROI convolution layer.

\noindent
\textbf{Loss function}
We use a multi-task loss that is similar to that defined in Mask RCNN \cite{maskrcnn}:
$L = L_{reg} + L_{cls} + L_{seg}$ where the bounding box regression loss $L_{reg}$ and the bounding box classification loss $L_{cls}$ are combined with the segmentation loss $L_{seg}$, which is the cross-entropy loss for all the pixels inside the ROIs.
Different from the loss in \cite{maskrcnn}, our loss is defined on the full image as opposed to on each ROI, which avoids considering overlapping pixels multiple times.


\section{Experiments}
\label{sec:results}
We demonstrate the performance of our method on segmentation of acetabulum bone from ultrasound images. 

\noindent
\textbf{Dataset}
Experiments were performed on a clinical dataset of 
3D ultrasound images of infant hips for diagnosis of developmental dysplasia of the hip (DDH). 
The segmentation of acetabulum is an essential step towards making diagnosis of DDH fully automated and less reliant on the operator. However the task is very challenging due to the highly noisy nature of ultrasound images, compared to CT or MRI images. Also the classes are highly imbalanced: the acetabulum only takes up a small portion of the image (0.3 percent in our
dataset).

Data collection was performed at the Radiology Department at University of Alberta.
The dataset consists of 49 three-dimensional ultrasound scans that has manually annotated segmentation labels for acetabulum provided by the lead pediatric musculoskeletal radiologist. We randomly divide them into a training set of 30 scans and a testing set of 19 scans. 
The scans from the same patient either all go into the training or testing set.
We crop some of the scans to make all the 2D slices share a unified size of 367x192 pixels (height by width). Most 3D volumes contain 256 slices.

\noindent
\textbf{Metrics}
We evaluated the segmentation on 2D slices using the following metrics: precision = TP/(TP+FP), recall = TP/(TP+FN), dice score = 2TP/(2TP+FP+FN), where TP, FP, FN denote True Positive, False Positive, and False Negative.
The results were the mean scores for all the 4864 test slices.

\noindent
\textbf{Implementation details}
We used the same VGG16-based RPN as in Faster R-CNN \cite{faster}, except that we reduced the padding of the first convolutional layer from 100 to 50.
The network was trained using SGD with a momentum of 0.9 and weight decay of 0.0005. Learning rate was set to 1e-5, and reduced by a ratio of 0.1 for every 50k iterations. 
We tried to keep the parameters the same as those in the original work to make a fair comparison.
The implementation was based on Caffe \cite{caffe}, where we added the implementation of ROI convolution layer. 
We compared our method with its FCN counterpart FCN-32s, and U-Net. We used the original Caffe implementation for FCN-32s. For U-Net, we tried three different implementations: the original Caffe implementation, an implementation from the 3D-U-Net \cite{unet3d}, and a Tensorflow \cite{tensorflow} implementation. The best result was reported. 

\noindent
\textbf{Results}
The test scores are summarized in Table \ref{tb:ds}. 
Note that the baseline method ``FCN-noPool5-scratch'' in the table represents the RPN-FCN network trained without the detection loss or the ROI convolution, i.e., with segmentation loss only.

\begin{figure} [hbtp]
	\begin{center}
        \includegraphics[trim=0 0.1cm 0 0.47cm,clip=True, width=0.69\linewidth]{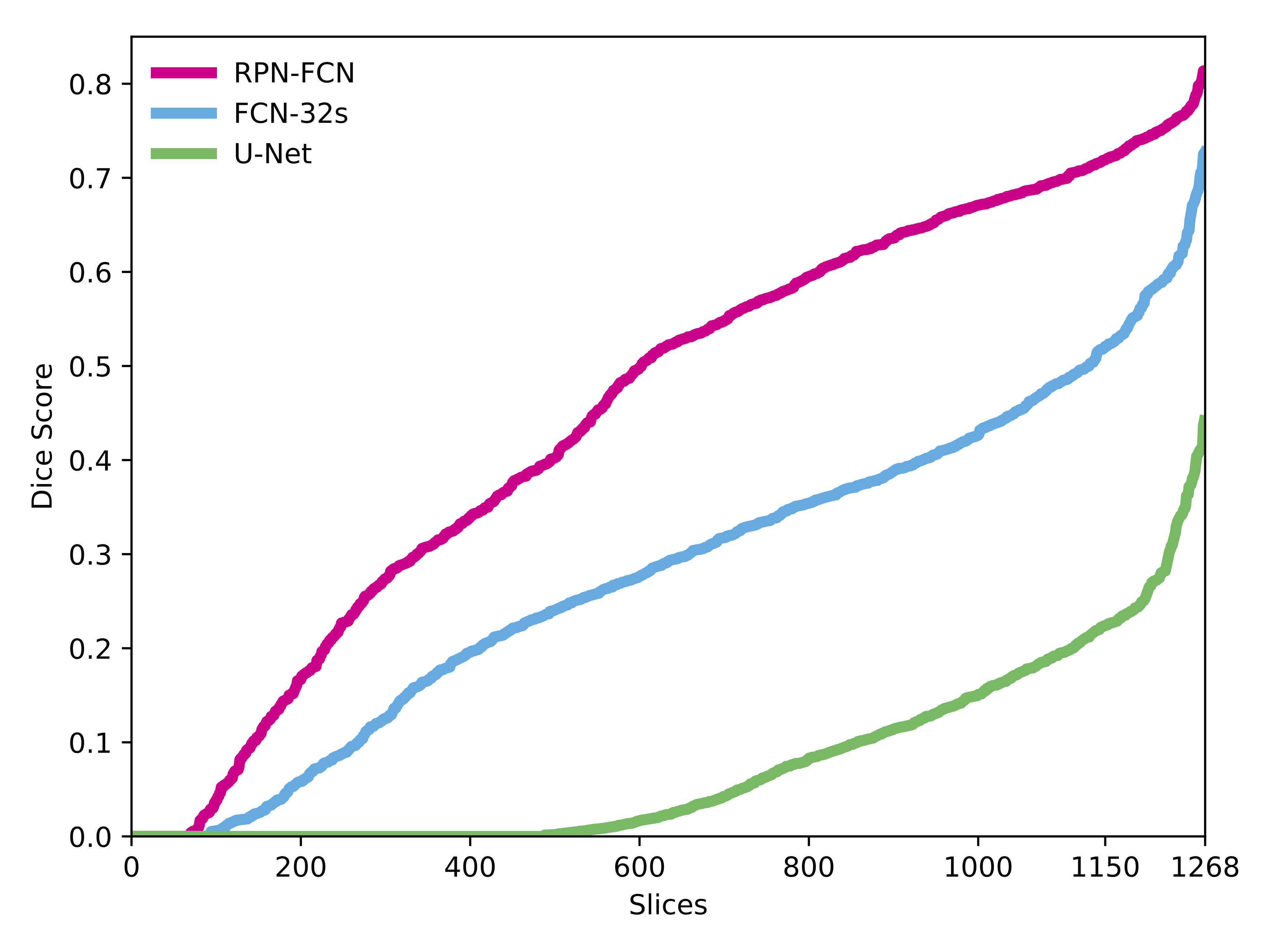}
        \vspace{-0.15in}
		\caption{Dice score of all the test slices that contain positive labels, sorted in ascending order of dice score.}
		\label{fig:sortdice}
	\end{center}
    \vspace{-0.25in}
\end{figure}

Our method RPN-FCN achieves better segmentation accuracy than the other methods. The distribution of the dice score are shown in Fig. \ref{fig:sortdice} by sorting the slices that contain positive labels based on their dice score. RPN-FCN yields better dice scores on a wide range of slices. Adding RPN on top of FCN-32s confines the segmentation problem in a smaller region. The result suggests that it is an effective way to train the downstream segmentation layers. 

\begin{table}[tbhp]
 \setlength{\abovecaptionskip}{-0.cm}
\setlength{\belowcaptionskip}{-0.cm}
\caption{Comparison of our proposed method \textbf{RPN-FCN} with FCN-32s, U-Net and the baseline method FCN-noPool5-scratch (RPN-FCN without using the detection loss or ROI convolution in training). Scores are averaged over all the test slices. \textbf{Bold} numbers denote the best of these methods. For all the methods, the best results from hyper-parameter tuning and different implementations are reported.}
\label{tb:ds}
\begin{center}
\vspace{-0.1in}
\begin{tabular}{|l|c|c|c|}
\hline
 Method & Precision & Recall & Dice Score \\
 \hline
 RPN-FCN         &  \textbf{0.297}   & \textbf{0.551} &  \textbf{0.386}  \\
 U-Net              		&  0.039   &  0.063 & 0.049 \\
 FCN-32s                   &  0.182  &  0.285 &  0.223  \\
 FCN-noPool5-scratch & 0.204 & 0.259 & 0.228 \\
 \hline
\end{tabular}
\vspace{-0.25in}
\end{center}
\end{table}
%

Our network is similar to FCN-32s in that the backbone is both VGG16 and the upsampling part is similar. The major difference lies in the use of ROI convolution after `conv5' layer of VGG16. Due to the removal of the pooling layer, the feature map has a higher resolution. It only needs to be upsampled by a factor of 16 instead of 32 as in the original FCN. 
Upsampling from a higher resolution feature map helps to retain the features of small objects. 

\begin{figure} [tbph]
	\begin{center}
	\hspace{0.1in} Me6401IM\_0063 \hspace{0.08in} Sc2201IM\_0022 \hspace{0.08in} So2601IM\_0022 \hspace{0.1in}
    
		\vspace{0.02in}
		\includegraphics[trim=0 1.5cm 0 3cm, clip=True,width=0.3\linewidth]{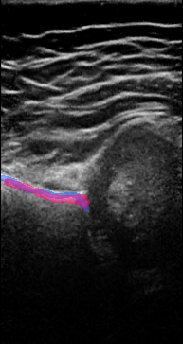}
		\includegraphics[trim=0 1.5cm 0 3cm, clip=True,width=0.3\linewidth]{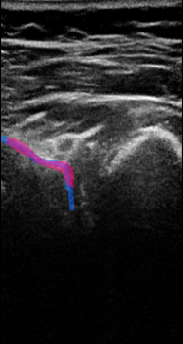}
		\includegraphics[trim=0 1.5cm 0 3cm,clip=True,width=0.3\linewidth]{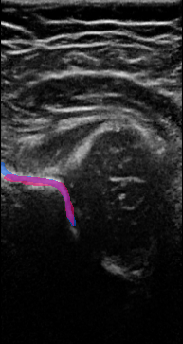}\\

		\vspace{0.01in}
		\includegraphics[trim=0 1.5cm 0 3cm, clip=True,width=0.3\linewidth]{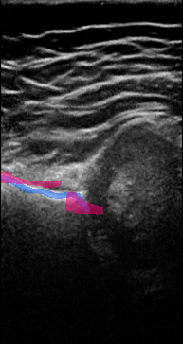}
		\includegraphics[trim=0 1.5cm 0 3cm, clip=True,width=0.3\linewidth]{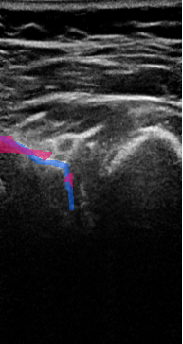}
		\includegraphics[trim=0 1.5cm 0 3cm, clip=True,width=0.3\linewidth]{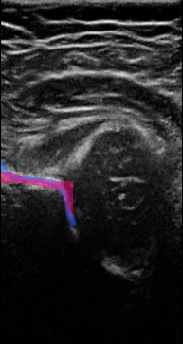}\\

		\vspace{0.01in}
		\includegraphics[trim=0 1.5cm 0 3cm, clip=True,width=0.3\linewidth]{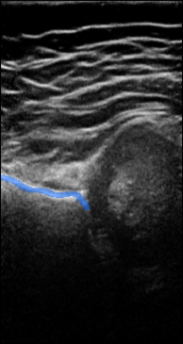}
		\includegraphics[trim=0 1.5cm 0 3cm, clip=True,width=0.3\linewidth]{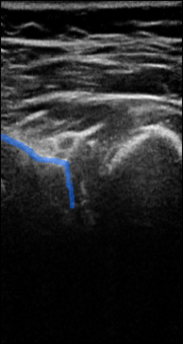}
		\includegraphics[trim=0 1.5cm 0 3cm, clip=True,width=0.3\linewidth]{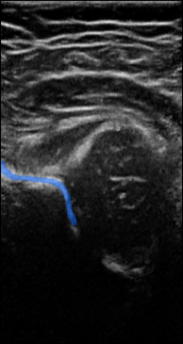}\\
		\vspace{-0.05in}
        \caption{Segmentation of slices on different volumes. Blue masks denote the ground truth; purple denotes the prediction. The heading in each column is the name of the volume containing the slice. The three rows correspond to the results of RPN-FCN, FCN-32s, U-Net, respectively.}
\vspace{-0.34in}
		\label{fig:vis}
	\end{center}
\end{figure}

The dice score is low compared to other medical image segmentation tasks. The main reason is that the object of interest is small. 
The representations learned by the network are dominated by the majority class: the background. U-Net introduced class weighting in the loss to combat this issue. We tried weighting by the class frequency but did not see improvement. The best result for U-Net was in fact obtained from the 3D U-Net implementation of the 2D network, where the input was a 2D image patch instead of the full image.
It might be possible to achieve better performance with U-Net by intensive parameter tuning on the class weights or the solver. But our main point here is that small object segmentation poses challenges on full-image-based methods. The detection mechanism in our proposed network offers an effective means for alleviating the issues in training the network.
We visualized the segmentation results of some slices in Fig. \ref{fig:vis}. There is no visible prediction for U-Net since it predicted all pixels as background. Our network was able to produce smooth regions when the shape is relatively clear in the view. 
%
%

\section{Conclusion}
\label{sec:conclusion}
\vspace{-0.1in}
We presented a novel method for improving the segmentation accuracy of fully convolutional networks.
Drawing inspiration from the object detection literature, we incorporated the Region-Proposal-Network (RPN) which was part of the backbone of Faster R-CNN. The bounding box prediction from RPN was used as an extra input to the downstream convolutional layers for segmentation to improve its localization capability.
We introduced a ROI convolution layer that directly convolved on all the ROIs in one pass without the need of ROI pooling or region-wise processing, and showed that this layer was trainable through backpropagation.
We essentially broke down the segmentation task into two steps: first localizing a rough region in which the object had been detected, then performing segmentation in the region. The proposed network was trained end-to-end on an ultrasound image segmentation task. We showed that by breaking down the segmentation problem into a joint detection and segmentation process, the segmentation accuracy was improved.

\noindent
\textbf{Acknowledgement} This work is supported by NSERC and Alberta Innovate scholarships. The authors would like to thank Abhilash Hareendranathan for insights and discussions. 
\bibliographystyle{IEEEbib}
\bibliography{myref}
\end{document}